# Event-driven dynamic trajectories reconstruction and measurement of mechanical parameters for fragments

Haoyang Li[a,b], Banglei Guan[b,*], Muxi Zha[b], Yifei Bian[b], Minzu Liang[c], Yang Shang[b], Qifeng Yu[b]

[a] *School of Aeronautics, Northwestern Polytechnical University, Xi'an, Shaanxi 710072, China*

[b] *College of Aerospace Science and Engineering, National University of Defense Technology, Changsha, Hunan 410073, China*

[c] *College of Science, National University of Defense Technology, Changsha, Hunan 410073, China*

## Abstract

During warhead detonation, high-density, high-speed, and mutually occluded fragments are generated. Their mechanical parameters (position, velocity, kinetic energy) directly determine the lethality of the warhead fragment field. However, high-intensity flash and smoke in detonation scenarios severely hinder the accurate acquisition of these mechanical parameters. To address this challenge, this paper integrates experimental mechanics approaches and presents an event-driven method for reconstructing the dynamic trajectories of fragments and measuring their mechanical parameters. As a novel brain-inspired visual sensor, event cameras offer microsecond-level temporal resolution and high dynamic range lighting change perception, overcoming the difficulty of accurately measuring high-speed targets under strong flash interference. The method constructs a multi-event-camera vision system, adopting three geometric constraints: time-correlated epipolar constraint to find potential matching event point pairs, and trifocal tensor line constraint plus local homography constraint to eliminate mismatches. A comprehensive probability model is established, with entropy weight method determining the weight of each constraint's probability to quantitatively filter mismatches. 3D trajectory reconstruction is achieved via spatial line-line intersection and nonlinear optimization. Finally, the velocity and kinetic energy of the fragments are calculated based on the reconstructed trajectory. Experimental results show the method ensures multi-fragment matching accuracy and supports visual quantitative analysis of matching probability. In simulations, the planar calibration board reconstruction errors are 0.041% and 0.128%, with 1.0% fragment velocity measurement error. In real experiments, UAV verification confirms trajectory reconstruction average relative error of 0.51%. This method provides reliable technical support for the mechanical damage evaluation of warhead fragment fields and the tactical protection design.

## 1. Introduction

The fragments generated by the explosion of a warhead are characterized by their small size, high speed, and large quantity. These fragments disperse rapidly after the explosion and pose a significant threat to surrounding targets at extremely high velocities[1-4]. The lethality effect of fragments is closely related to their motion, with typical motion parameters including trajectories and velocity distribution[5-8]. Therefore, comprehensively and accurately measuring and analyzing fragment motion parameters is of great significance for assessing the lethality field of the warhead and designing effective tactical protection[9-10].

Numerous studies on methods for testing fragment motion have been conducted. These methods can generally be categorized into contact-based techniques, such as target plate testing[11], and non-contact techniques, including light curtain testing methods[12], X-ray shadow imaging[13], radar velocity measurement[14], and high-speed camera testing[15]. Although each fragment measurement method can meet specific testing needs to some extent, they also have their own limitations. The target plate testing method, while simple and intuitive with the ability to obtain position information of the fragments, suffers from low efficiency, particularly during large-scale fragment testing[16]. The process of data acquisition and processing is time-consuming and requires considerable manual intervention[17]. Additionally, the contact between the fragments and the target plate may interfere with their normal dispersion paths, and errors in the angle of incidence could affect the accuracy of the measurements, making continuous testing difficult[18-19]. Radar testing utilizes the Doppler effect to achieve non-contact acquisition of fragment velocity information, making it suitable for hazardous environments[20-21]. However, its data analysis is complex, requiring advanced signal processing techniques, and it only provides velocity data without detailed descriptions of fragment position and shape. Therefore, radar testing often needs to be combined with other methods[22]. The light curtain testing method can precisely measure the flight paths and velocities of the fragments, especially for simultaneous multi-point measurements[23]. However, it struggles to accurately distinguish the positions of different fragments when multiple fragments cross the same beam simultaneously. Additionally, the target plates of the light curtain may be damaged in high-energy explosion environments, limiting its application in extreme conditions[24]. X-ray testing can avoid the interference of fire and smoke, enabling direct observation of fragment positions and motion trajectories[25-26]. However, it involves high equipment costs, complex systems, and demanding maintenance, with relatively low testing accuracy. Furthermore, due to the limitations in acquisition speed, it cannot continuously monitor the flight state of the fragments for extended periods[27-28].

Currently, research on fragment measurement typically employs high-speed cameras[29-32].

The high-speed photography testing method involves using two or more high-speed cameras to record images of fragment motion. By combining detection and tracking algorithms, the pixel coordinates of the fragments are obtained. Through matching algorithms, corresponding points in multiple views are identified. Subsequently, by conducting intersection measurements in three-dimensional space, the spatial coordinates of the fragments are calculated[33-35]. Zhou et al.[5] proposed a spatiotemporal distribution measurement method for fragments based on a high-speed camera network. This method uses the principle of binocular stereo vision, capturing image sequences of explosion fragments synchronously from different angles using four high-speed cameras. Li et al.[36-37] proposed a method for measuring the position of warhead fragments using two light-field cameras. Through the imaging principle of microlens arrays, they established a spatial position calculation model for the fragments and employed digital refocusing technology to address the issues of fragment occlusion and overlap. Hu et al.[38] proposed a method for tracking the motion trajectories and reconstructing the spatiotemporal distribution of warhead fragment motion based on high-speed stereo photography. This method aims to tackle the challenges of multi-object tracking and three-dimensional trajectory reconstruction for high-density fragment populations. Watson et al.[39] tracked the motion trajectories of fragments and measured their velocity in ultra-high-speed impact experiments using four-view segmented imaging technology. High-speed camera testing captures the motion trajectories of fragments and allows for detailed analysis, providing relatively precise measurements. However, due to the limitations of image sensors, it is difficult to achieve both high dynamic range and high temporal resolution simultaneously[40]. Additionally, measuring the initial speed of fragments can be challenging, and the sampling range of the dispersion area is also limited, affecting the comprehensiveness and accuracy of the measurements[41-42].

The event camera is a novel neuromorphic visual sensor, whose imaging principle differs from that of traditional optical sensors[43]. Conventional optical sensors record the grayscale value of each pixel at fixed time intervals, thereby obtaining synchronized image frames. In contrast, event cameras monitor the brightness changes of each pixel with extremely low latency. When the brightness change of a pixel reaches a threshold, an event containing timestamp, coordinate, and polarity information is triggered[44]. As a result, during the observation process, asynchronous event stream data are output[45-47]. Compared to traditional cameras, event cameras exhibit significant advantages in several aspects[48]. First, event cameras can respond to scene changes with extremely high temporal resolution, with output frequencies reaching up to 1 MHz, far surpassing the conventional frame-based cameras. This allows event cameras to capture high-speed moving targets without motion blur, enabling precise capture of rapidly changing details[49]. Additionally, event cameras utilize logarithmic response sensors, which offer a dynamic range of up to 140 dB, significantly higher than the 60 dB of traditional frame-based cameras. This enables event cameras to accurately capture dynamic information of targets under extreme lighting conditions, especially in environments where light and darkness coexist[43]. Due to their asynchronous output of event data, event cameras have response latencies on the order of microseconds, making them suitable for applications requiring rapid feedback with minimal time delay[50]. Finally, because event cameras only transmit brightness change data, they avoid the transmission of redundant data inherent in traditional frame cameras, thereby maintaining low power consumption[43]. Given these advantages, event cameras are highly suitable for scenarios involving high dynamic range and fast-moving targets[51]. Their application in moving targets measurement can overcome the limitations of optical metrology techniques based on image frames, providing reliable data support for moving target field measurements[52-54].

The traditional matching methods include SIFT feature point matching and template matching methods[55-56]. However, these methods are difficult to adapt to scenarios lacking texture features. Due to the sensitivity of event cameras to the edges of moving targets, they can clearly capture the motion trajectories of such targets. After selecting an appropriate time window, the motion trajectories often exhibit linear characteristics[57]. Based on this characteristic, the straight-line features of the motion trajectories of targets can be utilized for matching and reconstruction, enabling the recovery of target motion trajectories and the measurement of motion parameters[58-59]. At present, linear feature matching can be divided into two parts. One part of the studies focuses on the matching of photometric features (like grayscale and color) within the neighborhood of line segments[60-64]. Schmid et al.[65] proposed an algorithm that combines image grayscale information with multi-view geometric constraints to test line matching at both short and long distances. Bay et al.[66] used color histograms on both sides of the line to construct descriptors and perform initial matching, applying boundary constraints to filter out mismatches and iteratively obtain more matches. Wang et al.[67] established the MSLD descriptor, which exhibits strong robustness in line matching under conditions such as rotation, lighting variation, image blur, viewpoint changes, noise, JPEG compression, and partial occlusion. Zhang et al.[68] proposed a line matching algorithm that designed a new line descriptor (LBD) to describe the local appearance of lines and used the geometric constraints between line pairs for matching. Another part of the research involves linear feature matching based on geometric constraints of points or lines[69-73]. Jones[74] provided a detailed discussion on the line feature matching problem, summarizing existing solutions and proposing a framework for addressing line feature matching. Fan et al.[75] constructed affine invariants using the neighborhood points that are coplanar with the 3D lines. They calculated the similarity between lines by affine invariants. Lourakis et al.[76] proposed a unified method for matching coplanar points and lines in two views. This method employs a randomized search strategy, combining two lines and two points to construct a projection invariant, thereby obtaining a set of potentially matched points and lines. Subsequently, these candidate matches are verified through plane homography. Al-Shahri et al.[77] introduced a line matching algorithm based on epipolar geometry and coplanar constraints between line pairs. Testing showed that the proposed method outperformed the appearance-based and geometry-based methods in terms of matching accuracy. Chen et al.[78] addressed the issues of robustness and accuracy in segment matching under wide baseline conditions by hierarchically modeling the local structural features and geometric constraints. All the aforementioned methods are based on the research of static line feature matching in the form of traditional image frames, and there is still a lack of exploration of matching methods for moving targets based on line features.

This paper proposes a fragment trajectory measurement method using multiple event cameras based on geometric constraints. Firstly, a trinocular event camera measurement system is established to record the motion trajectories of targets. All potential matching trajectories are identified through the time correlation epipolar constraint, and the trifocal tensor line constraint and local homography constraint are adopted to filter all initially matched trajectories. Secondly, the matching results of the three types of geometric constraints are converted into probability forms, and a comprehensive matching probability model based on the entropy weight method is constructed to quantify the matching probability of each trajectory. Finally, the spatially intersected trajectories are further refined via a nonlinear optimization algorithm, and the average velocity and kinetic energy of each trajectory is calculated.

## 2. Method

This section includes the multi-view geometric constraint method, the comprehensive matching probability model based on the entropy weight method, and the method for 3D trajectory reconstruction and optimization. They focus on corresponding trajectory matching, quantitative analysis of matching accuracy, and target trajectory reconstruction respectively.

### 2.1 Multi-view geometric constraints

Here introduces three geometric constraints adopted for trajectory matching: all potential matching trajectories are determined via the time correlation epipolar geometry constraint, and all matching trajectories are verified by using the line constraint of the trifocal tensor and the local homography constraint, so as to accurately obtain the corresponding trajectory matching combinations among different event camera views.

**2.1.1 Time correlation epipolar geometry constraint.**

In the process of reconstructing the three-dimensional coordinates of a target from the two-dimensional image coordinates, it is first necessary to match the same target across different views to determine its imaging position in multiple field-of-view images. To achieve precise matching of multiple fragments, this paper proposes a stereo matching method based on time correlation epipolar constraint. When the target motion is recorded by binocular event cameras, due to the inherent delay in the event camera, the timestamps of ideal matching point pairs may exhibit slight differences. Therefore, a time difference threshold can be set to identify all potential matching event point sets[58, 79-81]. Within this set, epipolar constraints are then applied to filter out incorrect matches, ultimately obtaining accurate matching event point pairs. The specific formula for this process is shown below.

$$\begin{cases} |t_1 - t_2| \leq \Delta t \\ x' \mathbf{F} x \leq \varepsilon \end{cases} \tag{1}$$

In the formula, $t_1$ is the timestamp of an event in camera-1, $t_2$ is the timestamp of each event in camera-2, $\Delta t$ is the predefined time difference threshold, $x$ and $x'$ are the coordinates of the event points, $\mathbf{F}$ is the fundamental matrix, which is calculated from the camera's pre-calibrated internal and external parameters, and $\varepsilon$ is the residual threshold.

As shown in Fig. 1 (a) and (b), in the view $\mathbf{S}$, the timestamp information of a point $P_1$ to be matched is acquired. In the view $\mathbf{S}'$, all event points whose timestamps differ by $\Delta t$ from those of $P_1$ are identified. Then, based on the principles of epipolar geometry, the corresponding matching point in $\mathbf{S}'$ must lie on the epipolar line $l_2$. This allows the exclusion of event points in other trajectories that are close to the epipolar line $l_2$, thereby ensuring the matching accuracy of the event points. When $\Delta t$ is sufficiently small, the fragment targets in the compressed event frame images exhibit point features, and the corresponding points across views can be easily identified via the epipolar constraint.

After completing the matching of event points in the binocular view, RANSAC straight-line fitting is performed on the target trajectories composed of event point sets[82]. During the iteration process, a preset distance threshold from the point to the straight line is used to judge whether the point is an interior point of the fitted straight line. When the number of interior points exceeds a certain number, it indicates that there is a high consistency between the fitted straight line and the data point trajectory, thus the iteration is stopped and the parameters of the fitted straight line are output. By properly setting the fitting parameters, the straight-line trajectories of the target's motion can be accurately obtained. Finally, the straight-line trajectories are matched based on the number of corresponding matched event points on the lines.

### 2.1.2 Trifocal tensor line constraint

The motion trajectories of fragments typically exhibit linear features in event camera views, and the trifocal tensor constraint in three-view geometry can be utilized to match these linear trajectories. The advantage of three-view geometry over two-view geometry lies in its stronger constraint capabilities. In a three-view vision system, both points and lines in the three views can be constrained more rigorously, whereas in binocular vision, only points can be weakly constrained. The mathematical description of the constraint relationships in three-view geometry is called the trifocal tensor[83]. As shown in the Fig. 1(c), given a corresponding line in two images, this line can be uniquely determined in the third image. This transfer property is of great significance for establishing image correspondences between multiple views in stereo geometry[84].

In multi-target motion measurement scenarios, binocular vision suffers from structural limitations that lead to reduced matching accuracy and mismatches. To address this issue, We introduce the trifocal tensor line constraint on the basis of time correlation epipolar constraint to further screen the line combinations with known matching relations, thereby effectively eliminating incorrect matched lines. The trifocal tensor line constraint formula is as follows:

$$\mathbf{T}_i = a_i b_4^T - a_4 b_i^T \tag{2}$$

$$l^T = l'^T \begin{bmatrix} \mathbf{T}_1 & \mathbf{T}_2 & \mathbf{T}_3 \end{bmatrix} l'' \tag{3}$$

In the formula, $\mathbf{T}_i$ is the trifocal tensor, and $a_4$, $b_4$ represent the epipoles of the optical center of the first image in the second and third images, respectively. $a_i$ and $b_i$ are the $i$-th columns of the projection matrices of the second and third cameras. $l$, $l'$ , and $l''$ are the lines in the first, second, and third views, respectively.

The event stream information from the three views is compressed into the format of event frames. Under an appropriate time window, the trajectories of the moving targets in the event frame images are subjected to straight-line RANSAC fitting to obtain the linear parameters for each view. First, using the proposed trajectory matching method of the binocular stereo vision system, the trajectories in each pair of the three views are matched. The known linear trajectory matching relationships are then substituted into Equation (3) for verification, and any line combinations with incorrect matching relationships are identified.

### 2.1.3 Local homography constraint

When using feature point pairs to perform homography estimation, a key assumption must be satisfied: the selected feature point pairs for calculation must lie on the same plane[85]. Only when the feature points are coplanar can the projective transformation relationship described by the homography matrix hold accurately[86]. In line matching methods based on local homography estimation, an approximate processing approach is usually adopted, which assumes that points within a local region can be approximately considered coplanar[75, 87]. The optimal solution is determined by iteratively fitting the homography matrix multiple times, calculating its mapping error, and selecting the homography matrix with the smallest error. This method can tolerate the non-coplanarity of points to a certain extent; even if some points are not completely coplanar, a high-quality homography matrix can still be obtained through iterative optimization, provided that their geometric relationships within the local region are relatively consistent[88]. For fragments generated by an explosion, they can be approximately considered to originate from a single point and disperse, with the explosion center serving as the intersection point of the fragments' trajectories. Furthermore, for high-speed moving fragments, their motion trajectories can be approximately regarded as straight lines. Based on this, it can be assumed that: the straight-line trajectories of any two fragment targets lie on the same plane. Building on this assumption, local homography estimation can be performed on the trajectories of all fragment targets.

As shown in Fig. 1(d), the time correlation epipolar constraint is utilized to determine a pair of matching points $(\mathrm{p}_i, \mathrm{p}'_i)$ in the left and right views, where the homogeneous coordinates of the points in the left and right views are $\mathrm{p}_i = [x_i \quad y_i \quad 1]$ and $\mathrm{p}'_i = [x'_i \quad y'_i \quad 1]$, respectively. According to the projective transformation relationship between two planes, the two points satisfy the homography transformation relationship as shown in Equation (5), where **H** is the homography matrix.

$$\mathrm{p}'_i = \mathbf{H}\mathrm{p}_i \tag{4}$$

$$\begin{pmatrix} x'_i \\ y'_i \\ 1 \end{pmatrix} = \begin{pmatrix} h_{11} & h_{12} & h_{13} \\ h_{21} & h_{22} & h_{23} \\ h_{31} & h_{32} & h_{33} \end{pmatrix} \begin{pmatrix} x_i \\ y_i \\ 1 \end{pmatrix} \tag{5}$$

Equation (6) is rewritten into the form of a linear system of equations to obtain Equation (7). Here, Equation (7) can be expressed in the matrix form $\mathbf{Ah} = \mathbf{0}$, where h contains the parameter vector of the homography matrix H to be solved.

$$\begin{cases} h_{11}x_i + h_{12}y_i + h_{13} - x'_i h_{31} x_i - x'_i h_{32} y_i - x'_i h_{33} = 0 \\ h_{21}x_i + h_{22}y_i + h_{23} - y'_i h_{31} x_i - y'_i h_{32} y_i - y'_i h_{33} = 0 \end{cases} \tag{6}$$

$$\mathbf{A} = \begin{pmatrix} x_i & y_i & 1 & 0 & 0 & 0 & -x'_i x_i & -x'_i y_i & -x'_i \\ 0 & 0 & 0 & x_i & y_i & 1 & -y'_i x_i & -y'_i y_i & -y'_i \end{pmatrix} \tag{7}$$

$$\mathbf{h} = \begin{bmatrix} h_{11} & h_{12} & h_{13} & h_{21} & h_{22} & h_{23} & h_{31} & h_{32} & h_{33} \end{bmatrix}^T \tag{8}$$

Finally, the system of equations $\mathbf{Ah} = \mathbf{0}$ is solved by the least squares method. In addition, the last element $h_{33}$ of H is usually normalized to 1 to eliminate scale ambiguity, so at least four coplanar points are required to solve the homography matrix.

$$\min_{h} \| \mathbf{Ah} \|^2 \quad \text{s.t.} \quad \mathbf{h}^{\mathrm{T}}\mathbf{h} = 1 \tag{9}$$

The established homography matrix can be used to perform point transfer calculation between the left and right views via Formula (4), and then calculate the matching probability between trajectories. This method is elaborated in detail in Section 2.2.

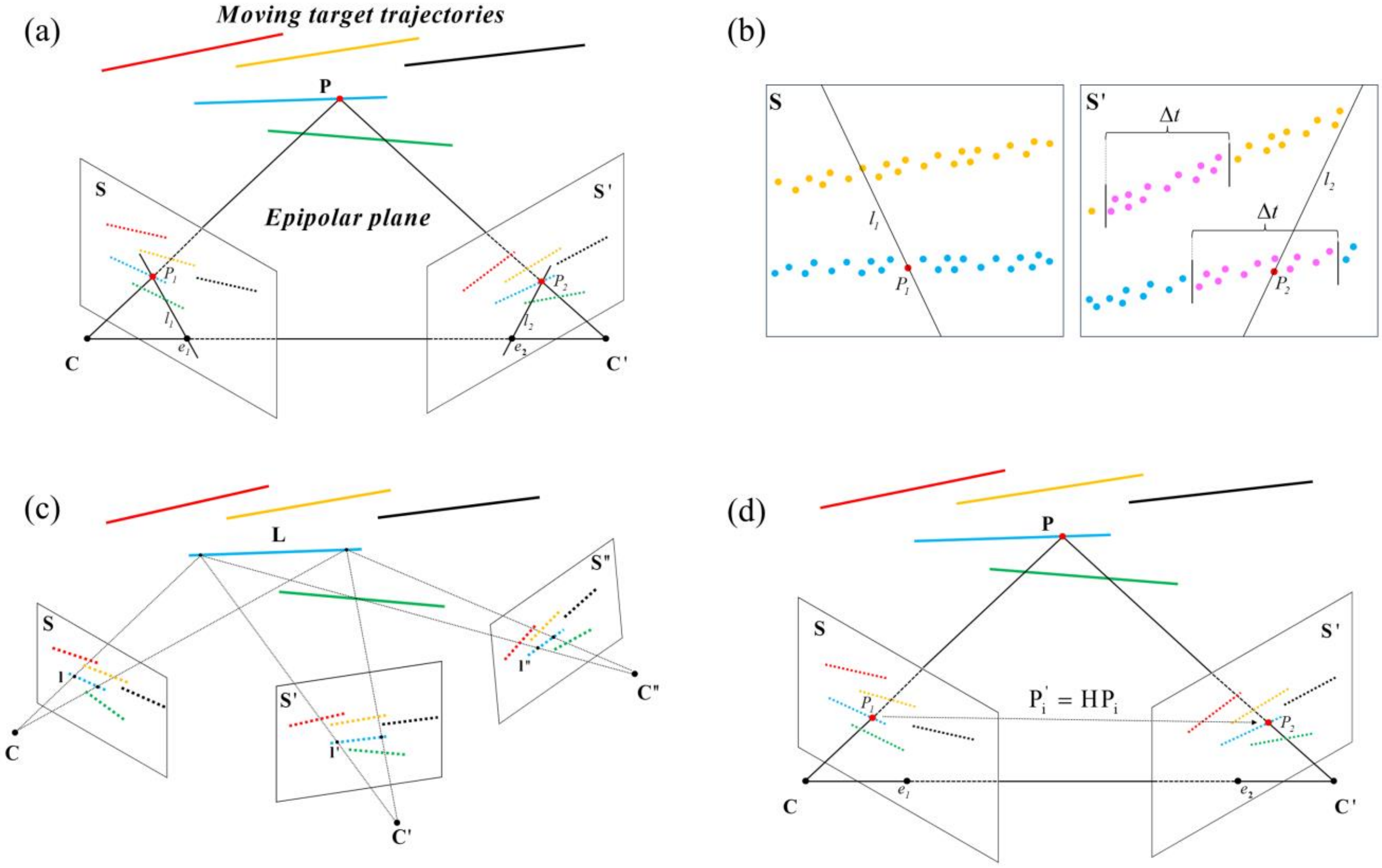

Fig. 1. Schematic diagram of multi-view geometric constraints: (a-b) Time correlation epipolar geometry constraint: Event points satisfying the epipolar relationship are searched within the time threshold;(c) Trifocal tensor line constraint: Matched lines in two views are transferred to other views;(d) Local homography constraint: Event points in one view are transferred to another view via the estimated local homography matrix.

### 2.2 Comprehensive probability model

Without background information, the matching relationships of a small number of motion trajectories in multi-view images can be easily determined by human visual inspection based on their relative positional relationships. However, when there are a large number of moving targets, it is difficult to identify corresponding trajectories through manual judgment. Furthermore, in real-world tests, the true value of dynamic trajectory matching for moving targets is also challenging to obtain. Thus, it is necessary to establish an evaluation model to quantitatively describe the trajectory matching accuracy. As shown in Fig. 2(a), for the time correlation epipolar geometry constraint, since a straight-line trajectory is composed of multiple event points, the greater the number of matching point pairs between two straight-line trajectories, the more likely these two straight-line trajectories are to be correctly matched. Therefore, the ratio of the number of matched event points on a straight-line trajectory to the total number of event points on that trajectory can be regarded as the matching probability of this straight-line trajectory. The formula is shown as follows:

$$p = \frac{N_{match}}{N_{total}} \tag{10}$$

where $N_{match}$ denotes the number of matched event points, and $N_{total}$ denotes the total number of event points on the trajectory.

As shown in the Fig. 2(b), after the initial matched trajectories are obtained via the time correlation epipolar constraint, the trifocal tensor constraint is adopted to screen the candidate matched trajectories. If there exists incorrect matching between trajectories, it will cause a significant discrepancy between the transferred line and the original corresponding line. Thus, this characteristic can be used to verify the matched trajectories.

As illustrated in the Fig. 2(c), any two pairs of matched straight-line trajectories are randomly selected in the camera view. The corresponding matched event point pairs on the selected trajectories are determined through the time correlation epipolar constraint. Taking a group of four pairs of matched points, we calculate the local homography matrix H using Equations (6)~(9). Subsequently, the points in the left view are transferred to the right view via the local homography matrix H. To further evaluate the quality of the estimated local homography matrix, the average distance from the four transferred points to the original matched points is calculated using Equation (11). A smaller average distance implies that the estimated local homography matrix is more accurate. If the two selected pairs of straight-line trajectories contain incorrect matches, it will lead to more than two straight-line trajectories actually participating in the estimation of the local homography matrix; that is, these trajectories will fail to satisfy the coplanarity assumption, ultimately resulting in the inaccuracy of the estimated local homography matrix, and the average distance from the transferred points of the trajectories to the original matched points will increase. Therefore, this principle can be utilized to judge the matched straight-line trajectories, thereby filtering out the incorrectly matched straight-line trajectories.

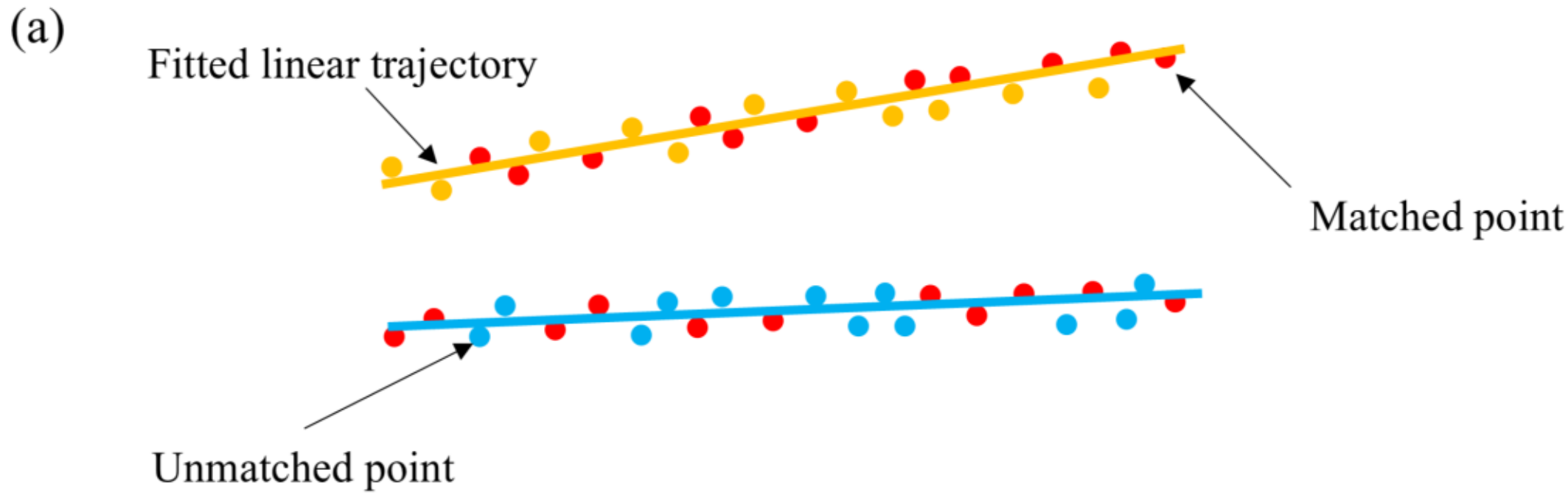

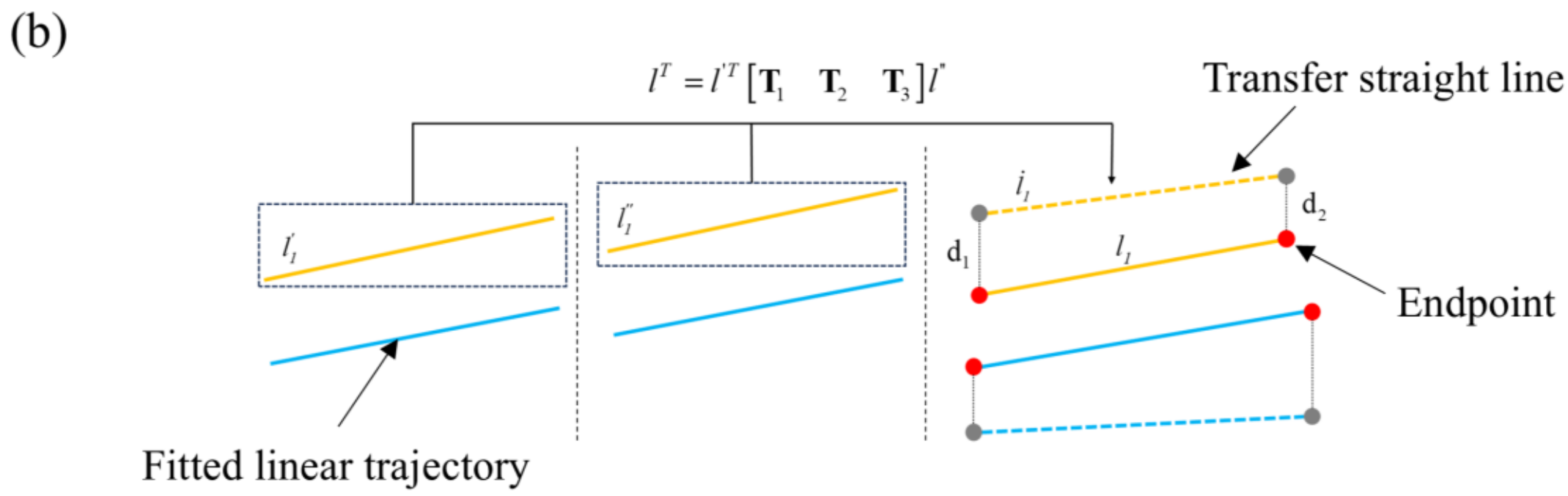

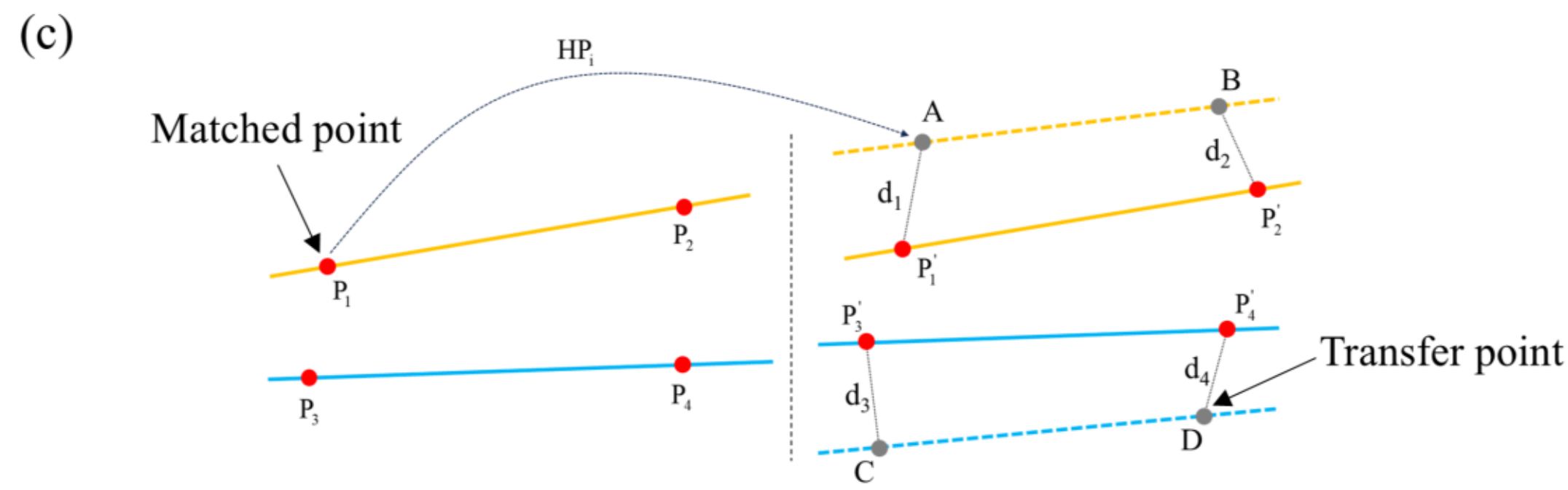

Fig. 2. Schematic diagram for constructing matching probabilities based on different geometric constraints: (a) Time correlation epipolar geometry constraint; (b) Trifocal tensor line constraint; (c) Local homography constraint.

For the trifocal tensor constraint and local homography constraint, these geometric constraint methods all involve transferring lines and points in one view to another view. The larger the distance between the transferred lines/points and the original lines/points, the higher the probability that the selected combination of straight-line trajectories is an incorrect match. Therefore, the distances of transferred lines and transferred points can be used as input parameters to construct a matching probability model via an exponential function. The model is expressed as the following formulas:

$$d_{\text{avg-1, avg-2}} = \sum_{i=1}^{n} d_i / n \tag{11}$$

$$p' = e^{-\lambda_1 \cdot d_{\text{avg-1}}}, p'' = e^{-\lambda_2 \cdot d_{\text{avg-2}}} \tag{12}$$

where $d_i$ denotes the point distance; $d_{avg-1}$ represents the average distance between the two endpoints under the trifocal tensor constraint; $d_{avg-2}$ denotes the average distance between the four endpoints under the local homography constraint; $\lambda_1$ and $\lambda_2$ are attenuation coefficients; and $p'$, $p''$ respectively represent the trajectory matching probabilities under the trifocal tensor constraint and local homography constraint.

To more accurately describe the matching results between straight-line trajectories in a quantitative manner, the trajectory matching results under each geometric constraint can be converted into probability forms. Then, reasonable weight coefficients are assigned to each probability to obtain the comprehensive probability that a pair of straight-line trajectories matches. When the comprehensive probability is lower than a certain threshold, the line matching combination is identified as incorrect. After eliminating the matching combinations with low comprehensive probabilities, the subsequent 3D reconstruction work is carried out, which ultimately ensures the accuracy of the constructed 3D straight-line trajectories. The comprehensive probability is calculated as shown in the following formula:

$$P_{i\text{-}match} = \alpha p_i + \beta p_i^{'} + \gamma p_i^{''} \tag{13}$$

In the formula, $\alpha$, $\beta$ and $\gamma$ denote the weight coefficients of the matching probabilities under the time correlation epipolar geometry constraint, the line constraint of the trifocal tensor, and the local homography constraint, respectively. $P_{i\text{-}match}$ denotes the comprehensive matching probability of trajectory.

To avoid the deviation caused by artificial subjective weight assignment for the weight coefficients in the formula, the entropy weight method can be adopted, i.e., a method that determines weight coefficients based on the objective characteristics of data[89]. As a key indicator for measuring system uncertainty, information entropy can intuitively reflect the dispersion characteristics of indicator data in the process of determining indicator weights. When there are significant differences in the observed values of a certain indicator, its information entropy value is small, which means the indicator contains more abundant effective discriminative information and should have a more prominent impact on the comprehensive evaluation results. Conversely, if the observed values of an indicator show a concentrated distribution with small differences, its information entropy value is large, indicating that the indicator has a lower information contribution and should have a relatively weaker impact on the evaluation results. The entropy weight method is precisely based on this characteristic: through the quantitative calculation of the information entropy of each indicator, it converts the objective dispersion degree of indicators into comparable weight coefficients. This weight determination method relies entirely on the statistical characteristics of the data itself, without introducing subjective judgments or empirical assumptions. Thereby, it effectively eliminates the interference of human factors on weight assignment, enabling the weight coefficients to truly reflect the objective importance of each indicator in the evaluation system[90].

First, the three matching probability indicators of all trajectories are standardized, where $x_i^{(j)}$ represents the standardized value of the *j*-th matching probability for the *i*-th trajectory, and $p_i^{(j)}$ denotes the matching probability of the *i*-th trajectory (*j*=1 corresponds to $p_i$, *j*=2 corresponds to $p_i^{'}$, and *j*=3 corresponds to $p_i^{''}$). Then, the proportion of each probability is calculated, with $p_{ij}$ being the proportion of the *i*-th trajectory in the *j*-th probability indicator, and m representing the total number of trajectories.

$$x_i^{(j)} = \frac{p_i^{(j)} - \min_i p_i^{(j)}}{\max_i p_i^{(j)} - \min_i p_i^{(j)}} \tag{14}$$

$$p_{ij} = \frac{x_i^{(j)}}{\sum_{i=1}^{m} x_i^{(j)}} \tag{15}$$

According to Formulas (16) and (17), the information entropy of each matching probability is calculated to measure the dispersion degree of each matching probability.

$$e_j = -\frac{1}{\ln m} \sum_{i=1}^{m} p_{ij} \ln p_{ij} \tag{16}$$

$$g_j = 1 - e_j \tag{17}$$

The weight coefficient of each matching probability is computed via the difference coefficient $g_j$.

$$\omega_j = \frac{g_j}{\sum_{j=1}^{n} g_j} = \frac{1 - e_j}{n - \sum_{j=1}^{n} e_j} \tag{18}$$

$$\alpha = \omega_1, \quad \beta = \omega_2, \quad \gamma = \omega_3 \tag{19}$$

Finally, the calculated weight coefficients are assigned to the matching probabilities under each constraint to obtain the comprehensive matching probability for each trajectory. Thus, the quantitative analysis of the matching accuracy of all trajectories is realized.

### 2.3 3D reconstruction of the event trajectory

As shown in Eqs. (20) and (21), $\mathbf{P}_j$ represents the projection matrix of camera-j , where $F_x^j$ and $F_y^j$ are the focal lengths and $C_x^j$ $C_y^j$ are the principal point coordinates of camera-j. $\mathbf{R}_j$ and $\mathbf{T}_j$ represent the rotations and translations between the first view and the *i*-th view, while $\tilde{x}_i^j$ and $\tilde{y}_i^j$ are the coordinates of points in the *j*-th view. $Z_c$ is the scaling factor. $X$ , $Y$, and $Z$ represent the three-dimensional coordinates of a point. The internal and external parameters of the camera can be obtained through the calibration procedure, and the projection matrix of each camera is calculated using Eq. (20). The central projection imaging relationship of the camera is shown in Eq. (21).

$$\mathbf{P}_j = \begin{bmatrix} F_x^j & 0 & C_x^j & 0 \\ 0 & F_y^j & C_y^j & 0 \\ 0 & 0 & 1 & 0 \end{bmatrix} \begin{bmatrix} \mathbf{R}_j & \mathbf{T}_j \\ \mathbf{0}^{\mathrm{T}} & 1 \end{bmatrix} = \begin{bmatrix} p_0^j & p_1^j & p_2^j & p_3^j \\ p_4^j & p_5^j & p_6^j & p_7^j \\ p_8^j & p_9^j & p_{10}^j & p_{11}^j \end{bmatrix} \tag{20}$$

$$Z_c \begin{bmatrix} \tilde{x}_i^j \\ \tilde{y}_i^j \\ 1 \end{bmatrix} = \mathbf{P}_j \begin{bmatrix} X \\ Y \\ Z \\ 1 \end{bmatrix} \tag{21}$$

Based on the known point matching and line matching combinations, the trajectories of the moving targets are reconstructed using the spatial line-line intersection method. As shown in Eq. (22), $x_i^j$ and $y_i^j$ represent the image coordinates after distortion correction. By combining the central projection equations of the two cameras, a linear equation system concerning the spatial point is obtained, thereby realizing the binocular intersection calculation of the spatial 3D point position. A point in a camera can generate two equations, and a pair of matched points in two cameras can generate four equations, forming a linear equation system. With four equations and three unknowns, this is an overdetermined system. Taking camera-1 and camera-2 as an example, the coordinates of the points are written in the form of column vectors, and the equation system can be rewritten in the form of Eq. (23). To solve for the three unknowns based on the four known equations, the least squares method can be used to obtain the 3D coordinates of the points.

$$\begin{cases} \left(x_i^j p_8^j - p_0^j\right) X + \left(x_i^j p_9^j - p_1^j\right) Y + \left(x_i^j p_{10}^j - p_2^j\right) Z + \left(x_i^j p_{11}^j - p_3^j\right) = 0 \\ \left(y_i^j p_8^j - p_4^j\right) X + \left(y_i^j p_9^j - p_5^j\right) Y + \left(y_i^j p_{10}^j - p_6^j\right) Z + \left(y_i^j p_{11}^j - p_7^j\right) = 0 \end{cases} \tag{22}$$

$$\mathbf{Ax} = \mathbf{b} \tag{23}$$

$$\mathbf{A} = \begin{bmatrix} x_i^1 p_8^1 - p_0^1 & x_i^1 p_9^1 - p_1^1 & x_i^1 p_{10}^1 - p_2^1 \\ y_i^1 p_8^1 - p_4^1 & y_i^1 p_9^1 - p_5^1 & y_i^1 p_{10}^1 - p_6^1 \\ x_i^2 p_8^2 - p_0^2 & x_i^2 p_9^2 - p_1^2 & x_i^2 p_{10}^2 - p_2^2 \\ y_i^2 p_8^2 - p_4^2 & y_i^2 p_9^2 - p_5^2 & y_i^2 p_{10}^2 - p_6^2 \end{bmatrix} \quad \mathbf{x} = \begin{bmatrix} X \\ Y \\ Z \end{bmatrix} \quad \mathbf{b} = \begin{bmatrix} p_3^1 - x_i^1 p_{11}^1 \\ p_7^1 - y_i^1 p_{11}^1 \\ p_3^2 - x_i^2 p_{11}^2 \\ p_7^2 - y_i^2 p_{11}^2 \end{bmatrix}$$

When a moving target is captured by binocular cameras in space, it forms corresponding trajectory information in the left and right images. By utilizing the calibrated intrinsic and extrinsic parameters of the binocular system and the correspondence of trajectory pixels, the 3D trajectory reconstruction of the target can be achieved using the triangulation algorithm. However, due to factors such as sensor noise, numerical errors in the algorithm, and other external disturbances, the reconstructed 3D trajectory may contain certain errors, making it difficult to achieve the desired accuracy. Therefore, to improve the accuracy of trajectory estimation, nonlinear optimization of the 3D trajectory is required. The goal of optimization is to minimize the reprojection error of the trajectory. Specifically, by iteratively adjusting the 3D motion trajectory of the target, the pixel distance between its reprojection on the binocular cameras and the actually observed image trajectory is minimized. For this purpose, the following objective function is defined:

$$\min_{X}\sum_{i=1}^{2}\sum_{j=1}^{N}\rho\left(\left\|\pi(K_i,d_i,R_i,t_i,P_{ij})-p_{ij}\right\|^2\right) \tag{24}$$

Where $X$ denotes the target 3D trajectory to be optimized, $i$ represents the left and right cameras of the binocular system, and $N$ stands for the number of trajectories. $K_i$, $d_i$, $R_i$, and $t_i$ respectively refer to the intrinsic parameter matrix, distortion coefficients, and pose parameters of the $i$-th camera. $P_{ij}$ denotes the 3D unknown parameters of the target, while $p_{ij}$ represents the image observation point corresponding to $P_{ij}$. $\pi$ represents the projection function and $\rho$ represents the robust loss function.

To solve this nonlinear optimization problem, the Levenberg-Marquardt algorithm is adopted. By combining gradient descent and the Gauss-Newton method, this algorithm can effectively find the global optimal solution among multiple local minima. The initial value is based on the initial 3D trajectory estimate obtained from triangulation, which provides a reasonable starting point for the optimization process, helping to accelerate the convergence speed and improve the stability of the optimization.

## 3. Numerical simulation and results analysis

Fragment scattering scenarios were simulated via Blender software in this section. The simulated moving targets were measured using the method proposed in Section 2, so as to verify the effectiveness and accuracy of the trajectory matching and 3D reconstruction methods.

### 3.1 Numerical simulation setup

As shown in the Fig. 3, Blender software is used to model the fragment motion trajectories. First, on the YZ plane, a trajectory path is established every 10° by rotating around the X-axis, with the rotation range from 10° to 80°, forming a fan-shaped plane containing 8 straight trajectory paths. Then, all trajectory paths within this fan-shaped plane are rotated around the Z-axis every 10° to generate a set of trajectory paths (with the rotation range from 0° to 40°), and finally 40 straight motion trajectory paths are obtained. A moving small ball target is set on each trajectory path, where the diameter of the small ball is 8 mm and the moving speed is 10 m/s. All small balls start moving from the initial point and spread along the trajectory paths. Three cameras are arranged in front of the trajectory paths, and each camera has the same internal parameters: a focal length of 35 mm, a resolution of 1280×720 pixels, and a frame rate of 500 fps. Under these settings, each camera view is rendered into a 500-frame video. As shown in Fig. 3(b), a frame of image is captured from the view of Camera 1, where each moving target is clearly visible, and the size of each target is approximately 4 pixels. As shown in Fig. 3(c), the RGB video is converted into event stream data using v2e, and the timestamp, coordinates, and polarity information of each target are obtained[91].

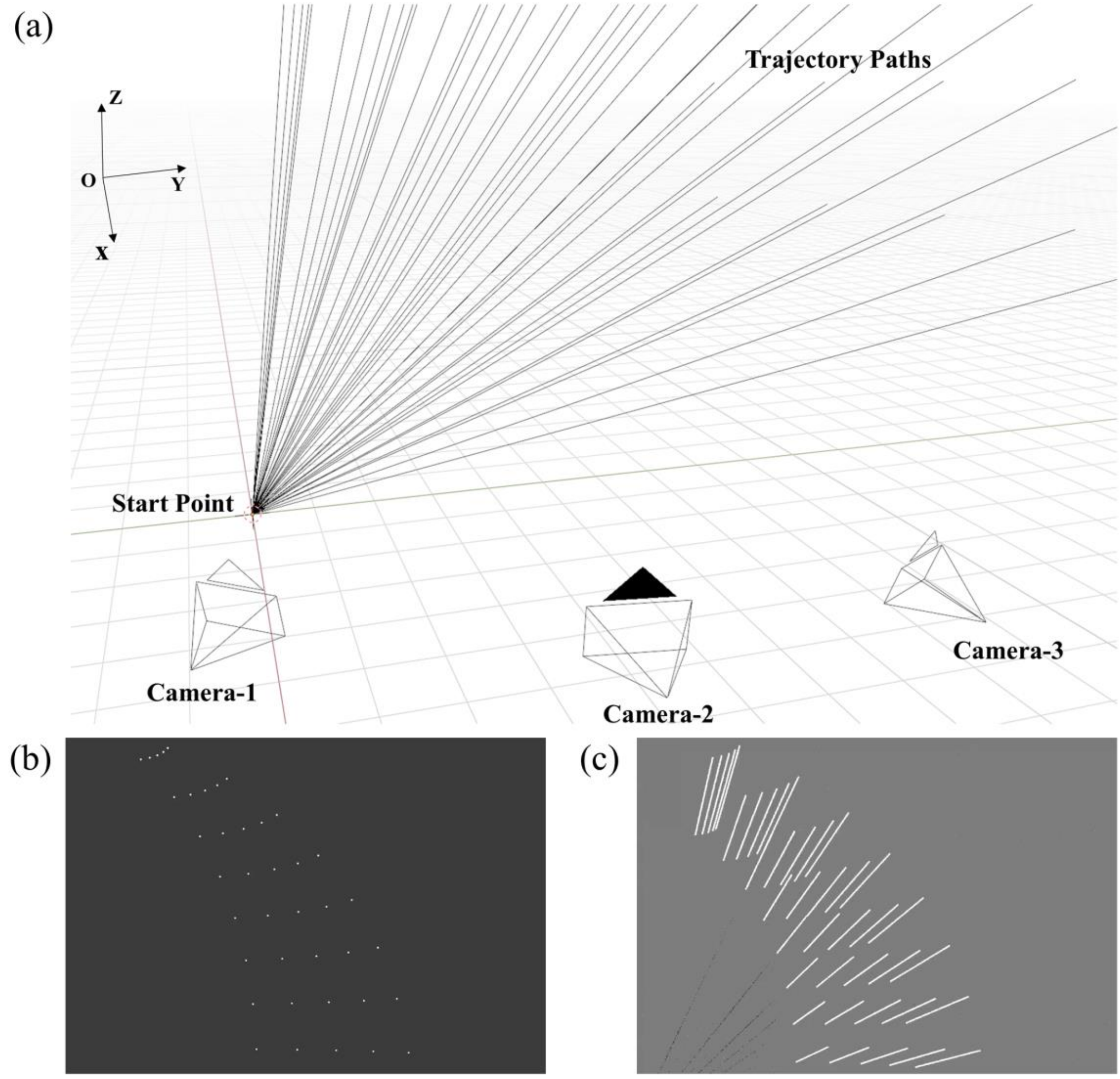

Fig. 3. (a) Numerical simulation experiment setup in Blender: Forty targets move uniformly along the trajectory path from the starting point, with three cameras arranged around to record the target motion trajectories. (b) Motion images of all targets under the perspective of Camera-1. (c) Event frame images obtained by processing Camera-1's images via v2e, with a time window of 150 ms.

### 3.2 Numerical simulation results analysis

The matching results of the simulated fragment motion trajectories captured by the Camera-1 and Camera-2 views are shown in Fig. 4(a) and Fig. 4(b). During the entire fragment dispersion simulation, a total of 40 event-stream straight-line trajectories were generated. First, the event-stream information of all straight-line trajectories was filtered using the time correlation epipolar constraint to obtain all candidate matched line combinations. Then, the straight-line trajectories recorded in Camera-3 were used to transfer the straight-line trajectories in Camera-2 and Camera-3 to the Camera-1 view via the trifocal tensor line constraint, thereby filtering out matched combinations with excessively large distance differences after line transfer. Furthermore, the local homography constraint was applied to all candidate matched line combinations to check and mark the matched combinations of straight-line trajectories with excessively large distance differences after point transfer. Notably, in the Camera-2 view, there exists a situation where a single straight line is matched with multiple straight lines in Camera-1 (e.g., matched trajectory numbers 1, 16, 21, and 29). This is because the positions of these targets are extremely close in the Camera-2 perspective, causing the targets to be identified as a single straight-line trajectory. For extremely dense targets, a single trajectory is used to approximate the motion trajectories of these targets. Therefore, if one view can distinguish the dense target trajectories, these dense targets can still be matched. Finally, as shown in Fig. 4(c), the comprehensive probability was applied to each pair of matched straight-line trajectories to quantitatively describe the accuracy of each matched combination. The probability weight of each constraint was determined using the entropy weight method. As shown in Fig. 4(d), 3D reconstruction was performed on the 40 matched straight-line trajectories; these trajectories were extended, and all extended lines approximately intersect at a single point, which is consistent with the actual trajectory modeling. The minimum distance between any two trajectories was calculated for all straight-line trajectories, and the average value was found to be 0.0155 m. As shown in Fig. 4(e), the average motion speed of each simulated fragment target was calculated and statistically analyzed: the average motion speed of all targets was 9.8450 m/s, with a relative error of 1.55% compared to the actually set speed of the targets.

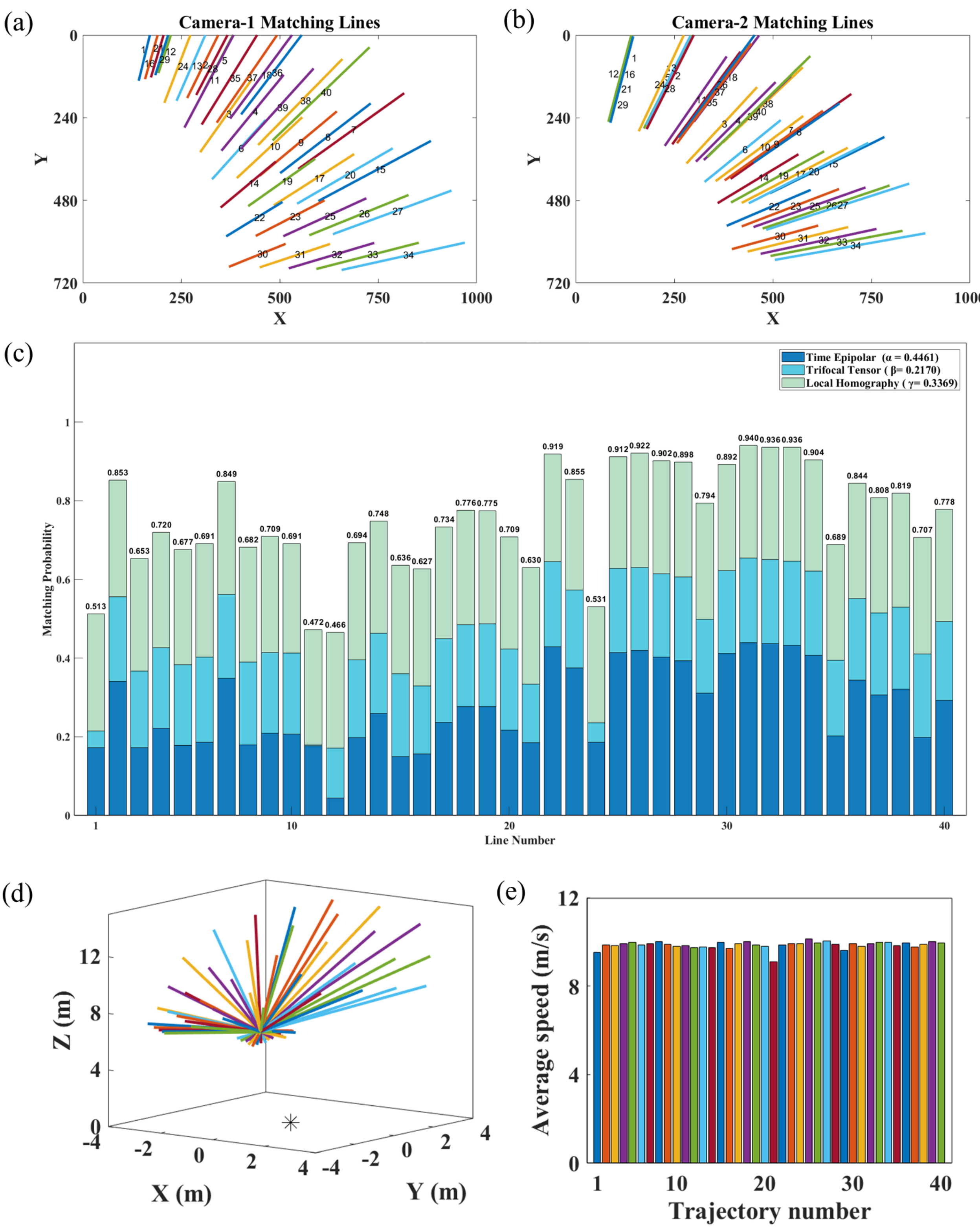

Fig. 4. (a-b) Trajectory matching results of Camera-1 and Camera-2: identical numbers in the two views indicate mutually matched trajectories. (c) Comprehensive matching probability results of all trajectories: the higher the probability, the greater the likelihood of being a correctly matched trajectory. (d) Extended line diagram of 40 trajectories after 3D reconstruction: asterisks indicate the position of Camera-1. (e) Statistical graph of the movement speeds of all targets.

## 4. Physical experiments and result analysis

In this section, the adaptability and robustness of the matching and reconstruction algorithms were further verified through physical experiments, and finally the 3D trajectory reconstruction accuracy and velocity measurement error were evaluated.

### 4.1 Simulated fragment measurement

In Section 4.1, launched steel balls were used as simulated fragments for experimental testing, verifying the performance of the proposed method in simulated fragment measurement under different conditions.

#### 4.1.1 Experiment setup

As illustrated in the Fig. 5, three event cameras are arranged on-site, with resolution of 1280×720 and an equivalent temporal resolution of 10k fps. By adjusting their positions appropriately, the three event cameras are configured to have a reasonable common field of view, with each camera equipped with a synchronization controller to ensure synchronized triggering. At a distance of 3~5 meters in front of the cameras, steel balls are used to simulate the motion of targets. The experiment is divided into few-target and multi-target scenarios. In the few-target experiment, a launcher was used to launch the targets, ensuring all targets were initiated from the same position. In the multi-target experiments, manual scattering was adopted; the initial movement positions of the targets were not confined to a single starting point, but instead, the targets were launched within an approximately acceptable range. The purpose of this experimental design is to further verify the adaptability of the non-homography target motion trajectory matching algorithm. For convenience, the few-target experiment is abbreviated as FTE (Few-Target Experiment), and the MTEs (Multi-Target Experiments) are abbreviated as MTE-1, MTE-2 respectively. The FTE and MTE-2 used steel balls with a diameter of 8mm, while the MTE-1 used steel balls with a diameter of 6mm.

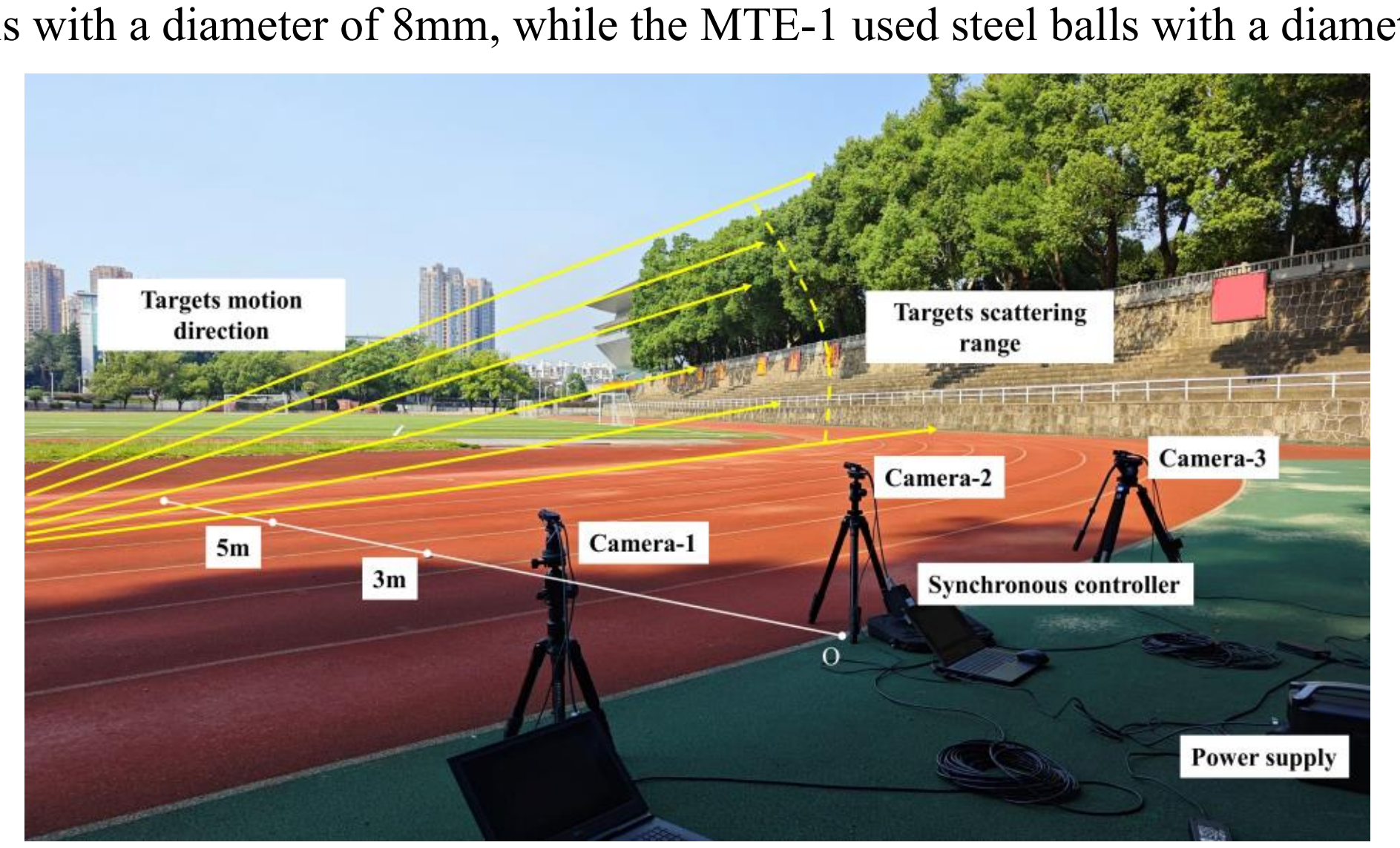

Fig. 5. Simulated fragment experiment: A trinocular event camera measurement system is established, with time synchronization between the cameras achieved via a synchronous controller. Simulated fragments are launched within the range of 3-5 m in front of the system, which cover the field of view of the cameras.

**4.1.2 Matching results and analysis**

As shown in Fig. 6($a_1$)~($a_3$), the time correlation epipolar constraint can accurately identify matching point pairs between the two views. After RANSAC line segment fitting is performed on the matched points, the target motion trajectories in the image coordinate system are obtained with high precision. Then, the line trajectories are matched based on the number of corresponding matched event points on the lines. Finally, the trifocal tensor line constraint and local homography constraint are used to further filter the already matched lines and eliminate incorrect matches. From the comprehensive matching probability results, it can be observed that the motion trajectories of the five targets are correctly matched. As shown in Figs. 6($b_1$)~($b_3$), in MTE-1, a total of 17 trajectories were effectively recorded, among which 16 were accurately matched. The matching precision is 100%, the recall rate is 94.1%, and the F1-score is 0.970. As also shown in Figs. 6($c_1$)~($c_3$), in MTE-2, a total of 19 trajectories were effectively recorded, among which 18 were accurately matched. The matching precision is 100%, the recall rate is 94.7%, and the F1-score is 0.973. To verify the ability of the comprehensive probability model to detect incorrectly matched trajectories, we artificially introduced incorrect matches of trajectories with close distances (trajectory numbers 17 and 18) into the matching results. The comprehensive probabilities of these two pairs of incorrectly matched trajectories were calculated as 0.057 and 0.101, respectively, which were the smallest among all matched trajectories. Therefore, the comprehensive matching probability model for straight-line trajectories constructed using the time correlation epipolar constraint, trifocal tensor line constraint, and local homography constraint can effectively detect incorrectly matched trajectories. In addition, in multi-target matching scenarios, although there are some trajectory overlaps, our algorithm can still achieve effective matching.

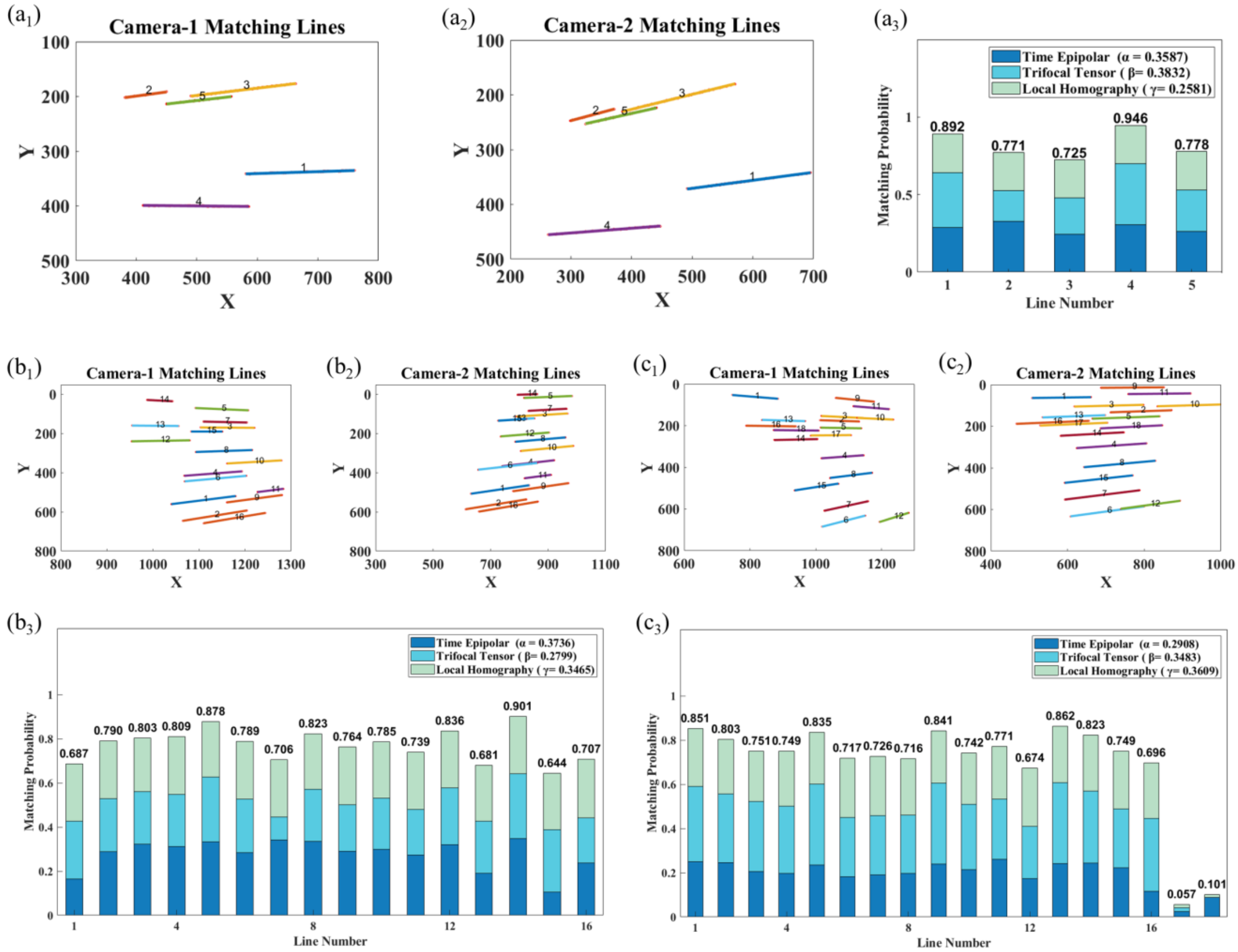

Fig. 6. Summary of Matching Results for Simulated Fragment Experiments: ($a_1$-$a_3$) Few-target Experiment (FEM): A total of 5 target motion trajectories are matched, and the comprehensive matching probability results indicate that all are correctly matched trajectories; ($b_1$-$b_3$) Multi-target Experiment (MTE-1): 16 trajectories are correctly matched; ($c_1$-$c_3$) Multi-target Experiment (MTE-2): 16 trajectories are correctly matched, and 2 incorrectly matched trajectories (No. 17 and No. 18) are detected.

#### 4.1.3 3D reconstruction and accuracy verification

As shown in Figs. 7(a)~(c), the 3D motion trajectories of the targets are reconstructed based on the known corresponding relationships of line matches. Red dots denote the 3D spatial coordinates of the target trajectories calculated via the two-view triangulation algorithm. The least squares method is used to fit the linear trajectories of the targets. Since the targets were simultaneously launched by the launcher, the backward extension of all trajectories approximately intersects at a single point. The average minimum distance between all 3D spatial trajectories is 0.0515 m, which further verifies the accuracy of trajectory matching and reconstruction. Since MTE-1 and MTE-2 adopted the manual throwing method, the backward extension of the trajectories does not intersect at a single point but forms a fan shape, which is consistent with the actual scenario. Steel balls with different diameters were used in the two experiments; accurate matching relationships and precise 3D motion trajectories were obtained through the comprehensive matching probability model and 3D reconstruction method. Therefore, the method proposed in this paper has certain applicability for fragments of different sizes.

To test the 3D reconstruction accuracy of target motion trajectories, a linear calibration board with 7×7 corners was measured, as shown in Fig. 7(d). The side length of each small grid is 100 mm. The 3D coordinates of the corners can be reconstructed via corner extraction and the binocular triangulation algorithm. By comparing the measurement results with the true values, the measurement error can be obtained. From the results, the average distance between adjacent points along the x-axis is 99.96 mm with a relative error of 0.041%, and the average distance between adjacent points along the y-axis is 99.87 mm with a relative error of 0.128%. The experimental results indicate that the reconstructed target trajectories have high measurement accuracy.

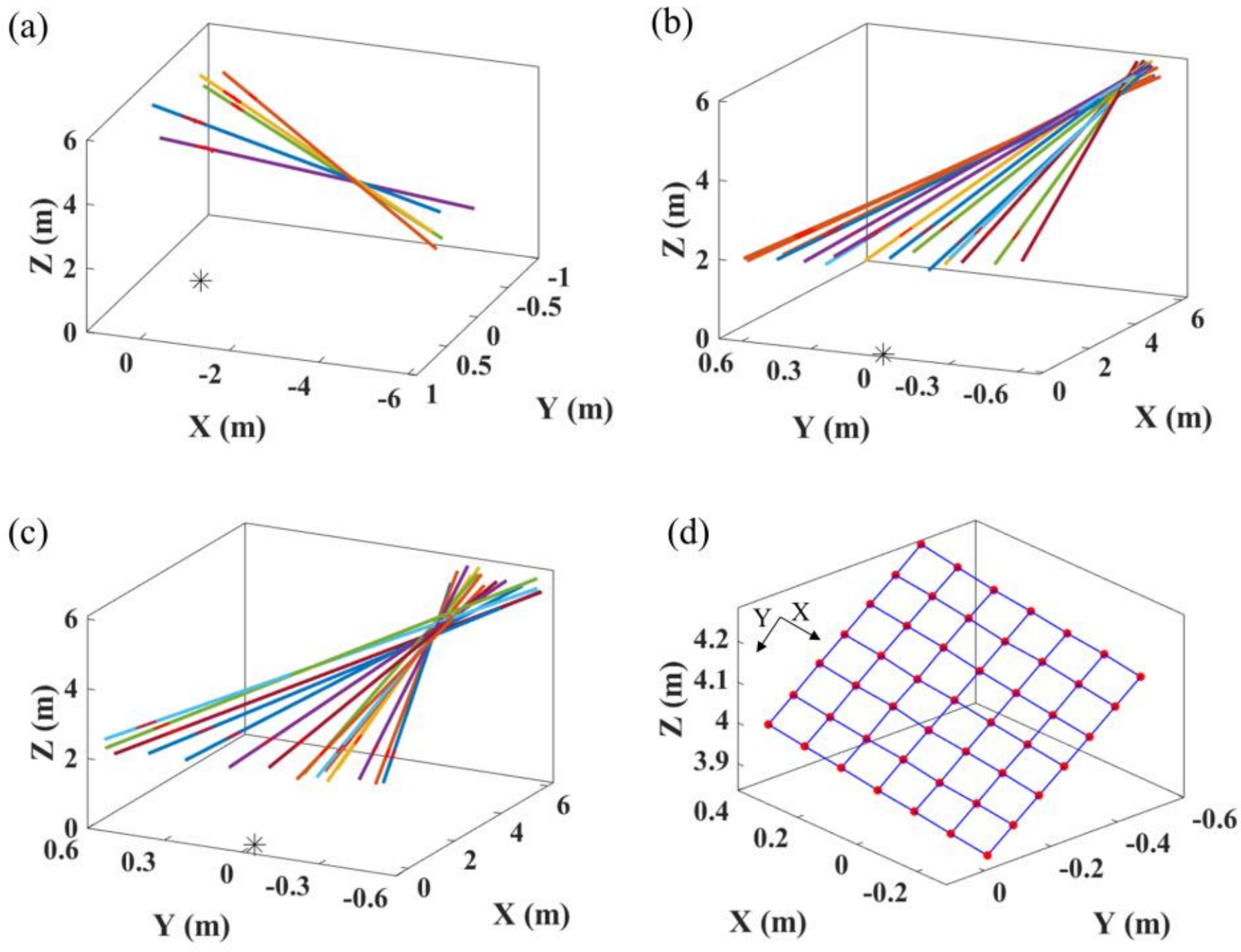

Fig. 7. Reconstruction trajectories with extension lines: (a) FTE; (b) MTE-1; (c) MTE-2. Asterisks indicate the position of Camera-1. (d) Verification results of 3D reconstruction accuracy: The reconstruction relative error in the x-direction is 0.041%, and that in the y-direction is 0.128%.

#### 4.1.4 Measurement of targets movement speed

Based on the 3D coordinates of trajectory endpoints and timestamp information, the average velocity of each trajectory is calculated. As shown in Fig. 8, in the FTE trajectory measurement experiment, the maximum average velocity of the targets is 27.8 m/s, the minimum velocity is 23.1 m/s, and the average velocity of the 5 targets is 25.4 m/s. In MTE-1, the maximum average velocity of the targets is 16.2 m/s, the minimum average velocity is 13.9 m/s, and the average velocity of the 16 targets is 15.1 m/s. In MTE-2, the maximum average velocity of the targets is 16.0 m/s, the minimum average velocity is 13.4 m/s, and the average velocity of the 16 targets is 14.6 m/s. The velocity difference between the two experiments is mainly attributed to the different target launching methods. Matching information of the entire trajectories can be obtained from the matched linear trajectories, thus enabling the reconstruction of the complete trajectories recorded in the views. The mass of the targets launched in each experiment is known. Based on this information, the average kinetic energy of the moving targets along the entire trajectories can be easily calculated. As shown in the line chart of Fig. 8, the total average kinetic energy of all trajectories in the FTE is 0.67 J, while those in MTE-1 and MTE-2 are 0.10 J and 0.21 J, respectively.

To further verify the accuracy of the measured target motion velocity, a high-speed launcher was used to launch a single target. The motion velocity of the single target was measured by an infrared light curtain velocimeter as the true velocity, which was then compared with the measured velocity recorded and calculated by the event camera. As shown in Fig. 9, the velocity accuracy verification test was repeated seven times, with an average relative error of the absolute value was 1.0%. On the basis of obtaining accurate matching correspondences and precise 3D trajectory reconstruction, the calculated average motion velocity of the targets also exhibits high accuracy.

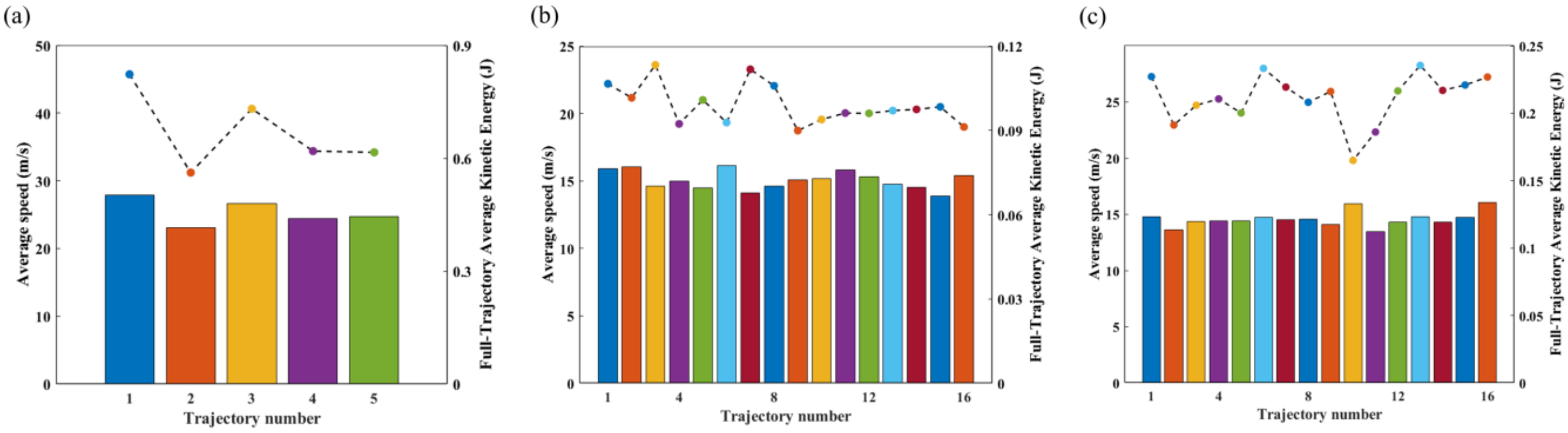

Fig. 8. Statistics of the average movement speed of targets: (a) FTE: The total average speed is 25.4 m/s, and the total average kinetic energy is 0.67 J; (b) MTE-1: The total average speed is 15.1 m/s, and the total average kinetic energy is 0.10 J; (c) MTE-2: The total average speed is 14.6 m/s, and the total average kinetic energy is 0.21 J.

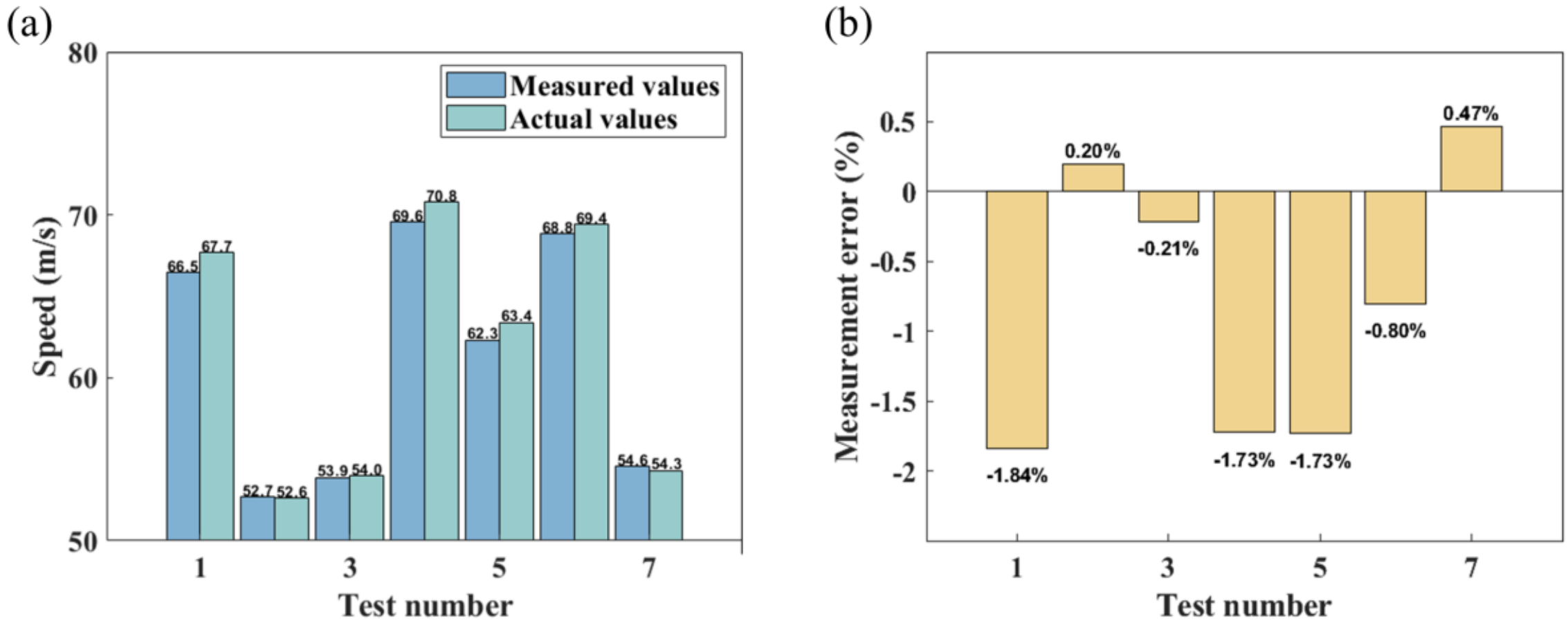

Fig. 9. Speed accuracy verification results: (a) Comparison between the measured speed and true speed of a single target; (b) Relative error of the measured speed of a single target.

### 4.2 Real fragment measurement

As shown in Fig. 10, an event camera measurement system was set up in the designated area. Due to the constraints of the experimental site, only a binocular event camera system was set up for local area fragments measurement in this experiment. The distance between the event cameras was approximately 340 meters, and each camera was equipped with a 400 mm telephoto lens to ensure the recording of the scattering process of local area fragments at a long distance. The event cameras were time-synchronized via a controller. Prior to the experiment, the event cameras were calibrated in advance using an unmanned aerial vehicle (UAV). The actual fragments were generated by the dynamic detonation of a certain type of missile, with the detonation center approximately 500 meters away from the cameras.

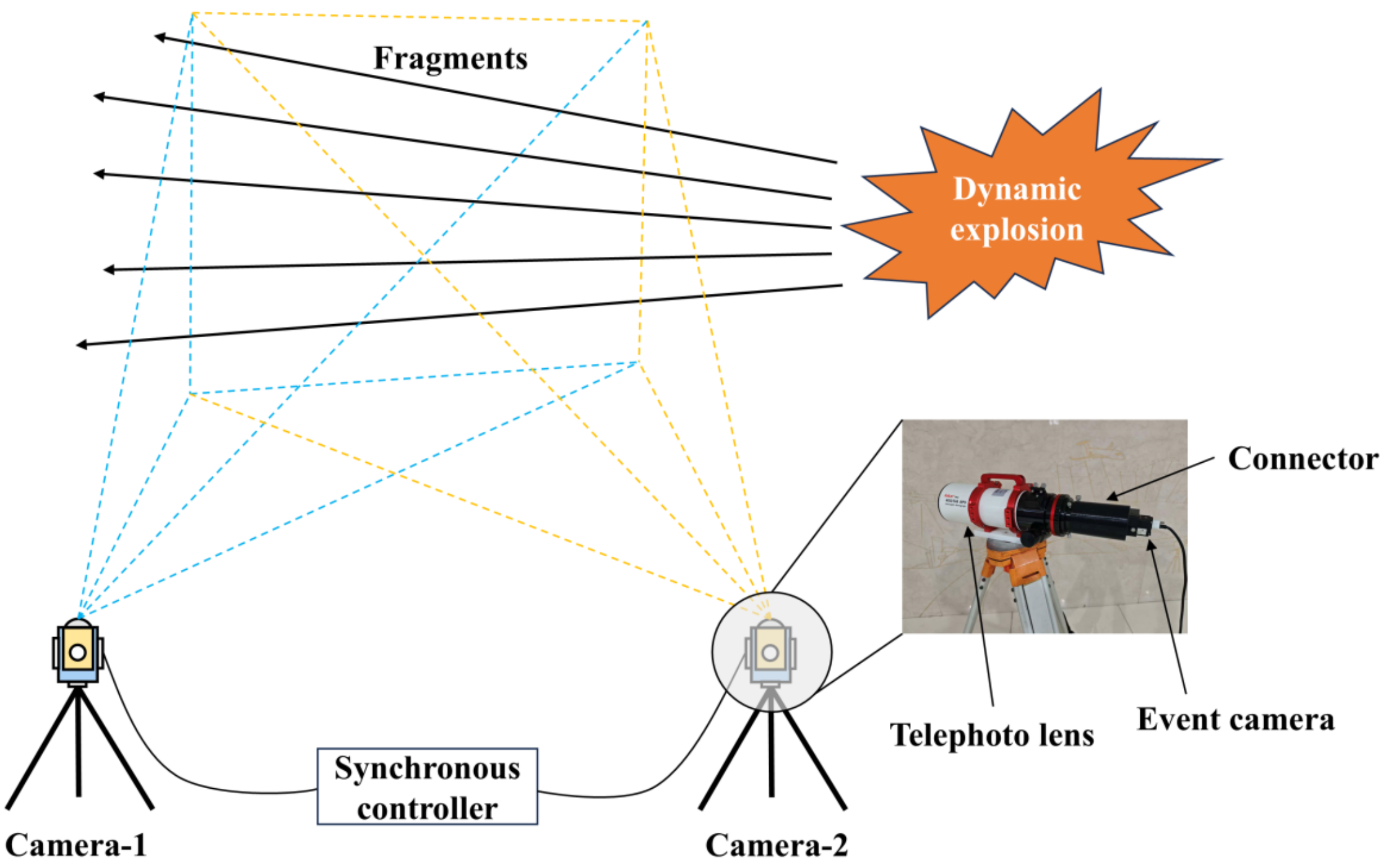

Fig. 10. Schematic diagram of the real fragment measurement experiment: Time synchronization of the binocular event cameras is achieved via a synchronous controller, with each event camera equipped with a 400 mm telephoto lens. The observed objects are fragments from the local position of missile dynamic explosion.

The actual fragment trajectory matching results are shown in Fig. 11(a-b). A total of 6 fragment trajectories within a specific local area were recorded. After matching via the time correlation epipolar geometry constraint and screening through the local homography constraint, 5 correct trajectory matching correspondences were finally obtained. Since two event cameras were used for measurements in the experiment, the trifocal tensor line constraint—intended for verifying matching relationships—was not applied. As shown in Fig. 11(c), the weight coefficient of the time correlation epipolar geometry constraint was 0.5758, and that of the local homography constraint was 0.4242. Except for Trajectory 5, the comprehensive matching probability of all other trajectories exceeded 50%. The comprehensive matching probability of Trajectory 5 was only 0.213, so it was regarded as an incorrect match and will not be considered in subsequent trajectory reconstruction and velocity measurement. Due to their dense distribution, Trajectories 2, 4, and 6 were identified as a single trajectory in Camera-2 and participated in the matching process. As exhibited in Fig. 11(d), the 5 fragment trajectories were subjected to 3D reconstruction and reverse extension; the reverse extension lines approximately intersect near a single region, which was consistent with the actual test conditions. To further verify the 3D reconstruction accuracy, the 3D coordinate points recorded by the differential GPS on the UAV were used to obtain the distances between the UAV's different hover positions in the field of view. By comparing these distances with those calculated via the intersection of the binocular event camera, the reconstruction error could be calculated. As shown in Fig. 11(e), 7 different hover positions were selected, and the average relative error of the distance between each position and the other positions was calculated. The final calculated average relative error of the UAV distance measurement was 0.51%. In Fig. 11(f), among the measured fragments, the maximum velocity was 363.2 m/s, the minimum velocity was 305.4 m/s, and the average velocity was 339.5 m/s. Based on the parameter information of preformed fragments, the average kinetic energy of each trajectory was calculated, and the total average kinetic energy of all trajectories is 121.7 J.

In summary, for actual fragment trajectories, the proposed matching, reconstruction, and measurement method can still ensure accuracy even when only two event cameras are used for measurement.

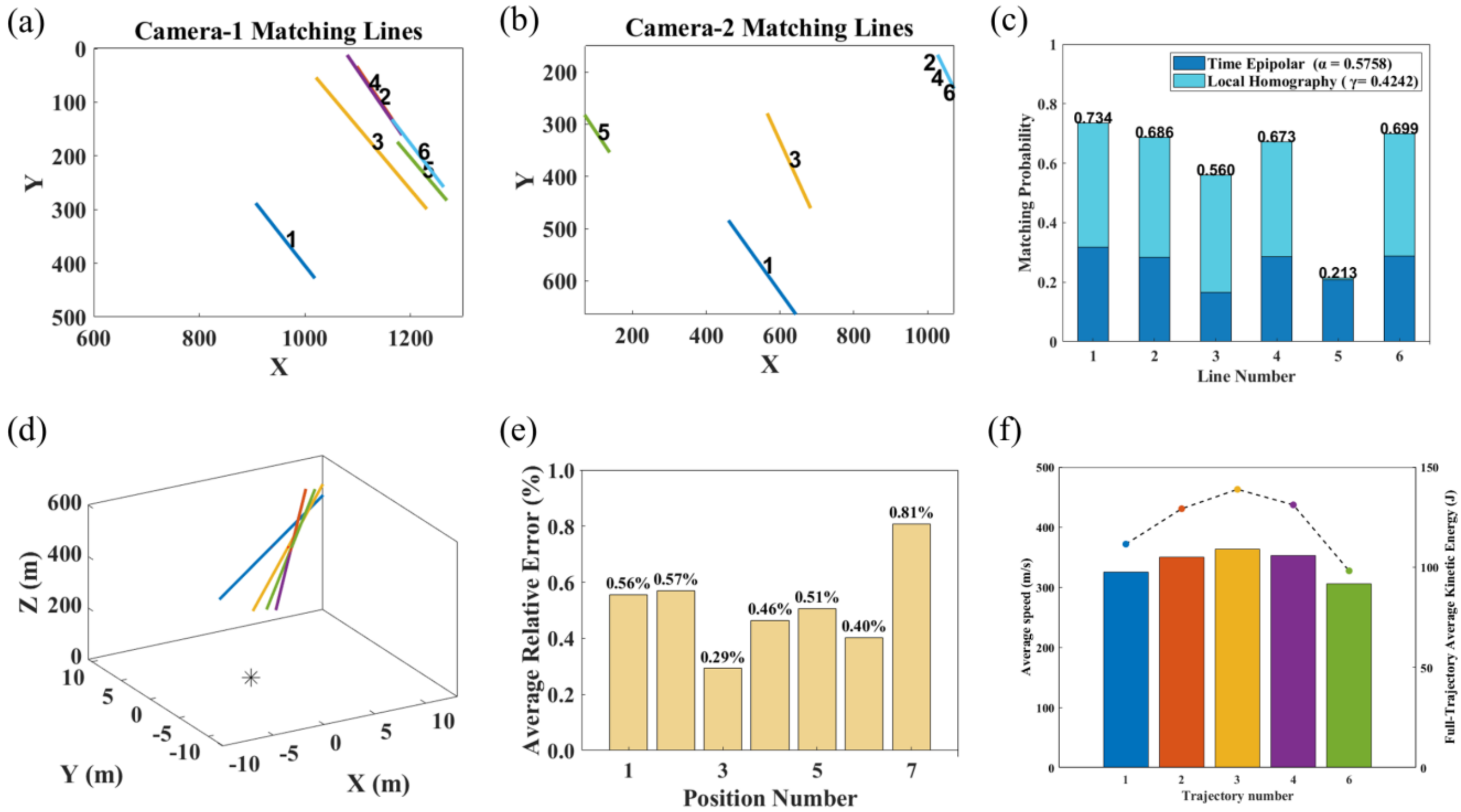

Fig. 11. Real fragment measurement results: (a-b) Trajectory matching results: A total of 6 fragment trajectories are matched; (c) Comprehensive matching probability results: Trajectory No. 5 is an incorrectly matched trajectory; (d) 3D reconstruction trajectories and extension lines: Asterisks indicate the position of Camera-1; (e) Verification results of 3D reconstruction accuracy: The relative error of trajectory 3D reconstruction is 0.51%; (f) Statistical results of fragment speed and kinetic energy: The total average speed is 339.5 m/s, and the total average kinetic energy is 121.7 J.

## 5. Conclusions

This paper proposes a motion target trajectory measurement method based on multi-view event cameras. The method constructs a multi-event-camera vision system, integrates time correlation epipolar constraint, trifocal tensor line constraint, and local homography constraint for trajectory matching, quantifies and evaluates matching results through a comprehensive probability model weighted by the entropy weight method. Finally, the three-dimensional reconstruction of the fragment trajectory and the measurement of speed and kinetic energy are achieved through spatial line-line intersection and nonlinear optimization. Through systematic verification by numerical simulations, simulated fragment experiments, and real fragment experiments, the effectiveness and reliability of the method are fully confirmed. The main conclusions of this paper are as follows:

1. The matching strategy integrating multiple geometric constraints and a comprehensive probability model can effectively solve the problem of dense fragment trajectory matching. The time correlation epipolar constraint screens potential matching point pairs through time thresholds and epipolar relationships; the trifocal tensor line constraint further eliminates mismatched lines by leveraging the strong constraint capability of three-view geometry; the local homography constraint verifies local geometric consistency based on the coplanarity assumption of fragment trajectories. Meanwhile, the comprehensive probability model combined with the entropy weight method converts the matching results of each constraint into objective and quantifiable probability indicators, enabling accurate identification of mismatched trajectories. Ensure the accuracy of the data sources used for subsequent mechanical parameter calculations.

2. The proposed method exhibits high measurement accuracy and strong robustness in both single-target and multi-target scenarios. In numerical simulations, the average distance between extended lines of fragment 3D trajectories is 0.0155 m, and the average relative error of velocity measurement is only 1.55%. In the simulated fragment experiments, the matching precision in the few-target scenario is nearly 100%, while the F1-scores in the multi-target scenarios reach 0.970 and 0.973, respectively; the relative errors of 3D reconstruction in the x-axis and y-axis are 0.041% and 0.128%, respectively; the average relative error of velocity measurement is as low as 1.0%. Additionally, the method can adapt to fragment overlapping scenarios and meet the measurement requirements for the mechanical parameters of fragments.

3. The technical advantages of event cameras are fully exerted in fragment trajectory measurement, and the method has clear engineering application value. In the real fragment measurement, five trajectories were successfully matched and reconstructed, and the fragment velocities and kinetic energy were measured. Meanwhile, the average relative error of trajectory reconstruction was verified to be 0.51% via the UAV. The proposed method can efficiently and accurately measure the trajectories and velocities of fragments. This can further provide reliable technical support for the evaluation of warhead fragment damage fields and the design of tactical protection.

## Acknowledgements

This work has been funded by the National Natural Science Foundation of China (Nos.12372189)